\documentclass[aps,pre,reprint,amsmath,amssymb]{revtex4-2}
\usepackage{txfonts}
\usepackage{algorithmic}
\usepackage{algorithm}
\usepackage{graphicx}
\usepackage{bm}
\usepackage{url}
\usepackage{color}
\usepackage{svg}

\begin{document}

\title{Integrated Utilization of Equations and Small Dataset in the Koopman Operator: \\Applications to Forward and Inverse Problems}

\author{Ichiro Ohta, Shota Koyanagi, Kayo Kinjo, and Jun Ohkubo}

\affiliation{Graduate School of Science and Engineering, Saitama University, Sakura, Saitama, 338--8570 Japan}

\begin{abstract}
In recent years, there has been a growing interest in data-driven approaches in physics, such as extended dynamic mode decomposition (EDMD). The EDMD algorithm focuses on nonlinear time-evolution systems, and the constructed Koopman matrix yields the next-time prediction with only linear matrix-product operations. Note that data-driven approaches generally require a large dataset. However, assume that one has some prior knowledge, even if it may be ambiguous. Then, one could achieve sufficient learning from only a small dataset by taking advantage of the prior knowledge. This paper yields methods for incorporating ambiguous prior knowledge into the EDMD algorithm. The ambiguous prior knowledge in this paper corresponds to the underlying time-evolution equations with unknown parameters. First, we apply the proposed method to forward problems, i.e., prediction tasks. Second, we propose a scheme to apply the proposed method to inverse problems, i.e., parameter estimation tasks. We demonstrate the learning with only a small dataset using guiding examples, i.e., the Duffing and the van der Pol systems.
\end{abstract}

\maketitle

\section{Introduction}
\label{introduction}

In physics, clarifying the underlying time-evolution equations in dynamical systems is one of the crucial tasks. On the other hand, data-driven approaches have attracted much attention in physics (for example, see Ref.~\cite{Brunton_book}). From a viewpoint of dynamical systems, several approaches based on the Koopman operator \cite{Koopman1931} have been proposed. In the Koopman operator approach, we consider a time-evolution for observable functions in a function space instead of a coordinate state space. Since the Koopman operator is linear, we can analyze nonlinear dynamical systems with linear algebra. There are several approaches to approximate the Koopman operator as a finite-dimensional matrix, such as the dynamic mode decomposition \cite{Schmid2010, Tu2014} and the extended dynamic mode decomposition (EDMD) \cite{Williams2015}; for details on the Koopman theory, see the review in Ref.~\cite{Brunton2022}. We here comment on the estimation of underlying time-evolution equations; the sparse identification of nonlinear dynamics (SINDy) \cite{Brunton2016} is the most famous data-driven discovery method for the underlying time-evolution equations, and the Koopman operator approach is also available for this aim, such as the generator EDMD (gEDMD) \cite{Klus2020} and the lifting technique \cite{Mauroy2020}. Although the main topic here is not the estimation of time-evolution equations, knowledge of the time-evolution equations plays an important role in our present work, as we will see below.

One of the disadvantages of the data-driven approach is the necessity of large datasets; it is well known that machine learning models, especially deep neural networks, typically need large amounts of data. Therefore, there are many works to improve accuracy with limited datasets by using prior knowledge. For example, some works combine energy conservation laws with machine learning methods \cite{Greydanus2019,Mattheakis2020,Zhang2024}. Some works focus on so-called physics-informed machine learning, which aims to treat partial differential equations in combination with prior knowledge \cite{Raissi2019}; see the review paper in Ref.~\cite{Karniadakis2021}.

There are various cases in which we have prior knowledge. One of the common cases is that one knows the time-evolution equations, but the values of the coefficient parameters in the time-evolution equations are unknown; this situation is common in data assimilation tasks \cite{Evensen_book}. We here refer to this situation as \textit{ambiguous prior knowledge}. As far as we know, there is no work to combine the Koopman operator approach with the ambiguous prior knowledge. Since the Koopman operator approach is suitable for linear analysis, it is preferable from the physics viewpoint to employ the ambiguous prior knowledge to achieve learning with a small dataset. Then, we pose the following questions:
\begin{itemize}
\item (forward problem) Is it possible to exploit the ambiguous prior knowledge to construct the Koopman matrix with a small dataset?
\item (inverse problem) Is it possible to estimate the unknown parameters in the time-evolution equations using the Koopman operator approach from only a small dataset?
\end{itemize}

In the present paper, we propose methods to integrate the ambiguous prior knowledge with the EDMD algorithm. Here, the ambiguous prior knowledge corresponds to the situation where we know the time-evolution equations, but the parameter values are unknown, as stated above. To combine the ambiguous prior knowledge with the Koopman matrix, we employ the duality relations \cite{Ohkubo2022}; a Koopman matrix is constructed from the underlying time-evolution equations with randomly selected parameters. Then, the constructed Koopman matrix is updated with the aid of the online EDMD algorithm \cite{Zhang2019}. We will also discuss an initial matrix for the online EDMD; in the conventional online EDMD, a known dataset generally leads to the initial matrix, but we cannot employ the conventional approach because of the small size of the datasets. For the forward problem, the proposed method enhances the accuracy only with a small dataset. Extending the proposed method, we address the inverse problem to estimate the unknown parameters, achieving reliable identification with only 10 snapshot pairs. We give numerical experiments for the forward and inverse problems using guiding examples, i.e., the Duffing and the van der Pol systems.

The remainder of this paper is composed as follows. Section~2 gives a brief review of the Koopman operator approach. Section~3 yields the first contribution of the present paper; we propose a method to combine the duality relation and the online EDMD. Numerical demonstrations are also given in Sect.~3. Section~4 is the second contribution, in which a scheme for the inverse problem is proposed. We give a summary and comments on future work in Sect.~5.

\section{Brief Review of Previous Works}
\label{sec_review_Koopman}  

\subsection{Snapshot pairs}

In the Koopman operator approach, it is common to consider a dataset obtained from discrete-time observations. Let $\bm{x}_{i} \in \mathbb{R}^{D}$ be a vector at discrete time $i$ in a $D$-dimensional state space. Assuming a discrete time-evolution function of a target nonlinear dynamical system, $F: \mathbb{R}^{D} \to \mathbb{R}^{D}$, we have
\begin{align}  
\bm{y}_{i} \equiv \bm{x}_{i+1} = F(\bm{x}_{i}),
\label{eq_discrete_evolution}
\end{align}  
where we define the state vector after the discrete time-evolution as $\bm{y}_{i} \in \mathbb{R}^{D}$. 

In this work, we assume that there are underlying continuous time-evolution equations that lead to the discrete dynamical system in Eq.~\eqref{eq_discrete_evolution}:
\begin{align}
\frac{d}{dt} \bm{x}(t) = \bm{f}(\bm{x}(t); \bm{\theta}),
\label{eq_time_evolution}
\end{align}
where $\bm{f}: \mathbb{R}^{D} \to \mathbb{R}^{D}$ is the generator of the time-evolution, and $\bm{\theta}$ is the parameter set. For example, if we set $\bm{x}_i = \bm{x}(t)$, it is possible to connect the discrete time-evolution in Eq.~\eqref{eq_discrete_evolution} with the continuous time-evolution by taking $\bm{y}_{i} = \bm{x}(t+\Delta t_{\mathrm{obs}})$, where $\Delta t_{\mathrm{obs}}$ is the time-interval for the observation. Although the original EDMD algorithm, as we will review below, does not necessarily need these underlying equations, we will use the underlying time-evolution equations \textit{with unknown parameter values}, $\bm{\theta}$, as the prior knowledge in the following sections.

A pair of vectors before and after time-evolution, $(\bm{x}_i, \bm{y}_i)$, is called a \textit{snapshot pair}. When there are $m$ snapshot pairs, the alignments of these vectors lead to the following matrices:
\begin{align}  
X_{m} =[\bm{x}_{1}\ \bm{x}_{2}\ \dots\ \bm{x}_{m} ], \quad
Y_{m} =[\bm{y}_{1}\ \bm{y}_{2}\ \dots\ \bm{y}_{m} ].
\label{eq_data_matrix}
\end{align}

\subsection{Koopman operator, dictionary, and EDMD}

The Koopman operator is an infinite-dimensional linear operator that yields the time-evolution of an observable function $\phi(\bm{x}): \mathbb{R}^{D} \to \mathbb{C}$. The time-evolution in the observable function space is governed by an operator defined as
\begin{align}
\mathcal{K} \phi = \phi \circ F.
\label{eq_Koopman_op}
\end{align} 
The operator $\mathcal{K}$ is known as the Koopman operator. Using the Koopman operator $\mathcal{K}$, we have
\begin{align}
\mathcal{K} \phi(\bm{x}_{i}) = \phi \circ F(\bm{x}_{i}) = \phi(\bm{y}_{i}).
\label{eq_Koopman_op_action}
\end{align}  
In Eq.~\eqref{eq_Koopman_op}, the left-hand side represents the action of the Koopman operator $\mathcal{K}$ on the observable function $\phi$, while the right-hand side represents the composition of the observable function $\phi$ with the time-evolution function $F$, illustrating the evolution of observable functions in the infinite-dimensional space. Equation~\eqref{eq_Koopman_op_action} means the following fact: due to the time-evolution of the observable function, one can evaluate the value of the observable function at the coordinates $\bm{y}_i$ after time-evolution, $\phi(\bm{y}_i)$, by substituting the coordinates $\bm{x}_i$ for the function $\mathcal{K}\phi$.

As denoted above, the Koopman operator is infinite-dimensional because $\mathcal{K}$ acts on elements in the function space. Hence, $\mathcal{K}$ should be approximated in a finite-dimensional vector space spanned by a set of basis functions. The set of basis functions is called a dictionary, which leads to an approximated finite-dimensional matrix $K$. Let $N_{\mathrm{dic}}$ denote the number of dictionary functions. Then, the dictionary is written as
\begin{align}
\bm{\psi} (\bm{x}) = 
\begin{bmatrix}
\psi_1(\bm{x}) &
\psi_2(\bm{x}) &
\cdots &
\psi_{N_{\mathrm{dic}}}(\bm{x}) 
\end{bmatrix}^{\top},
\label{eq_dictionary}
\end{align}
where $\psi_n$ is the $n$-th dictionary function.  

Here, assume that the observable function $\phi(\bm{x})$ is expressed using the dictionary functions as follows:
\begin{align}
\phi(\bm{x}) = \sum_{n=1}^{N_{\mathrm{dic}}} a_{n}^{\phi} \psi_{n}(\bm{x}),
\label{eq_observable_basis_expansion}
\end{align}
where $\{a_{n}^{\phi}\}_{n=1}^{N_{\mathrm{dic}}}$ represents the coefficients for the expansion of the observable function in terms of the dictionary functions. Although Eq.~\eqref{eq_observable_basis_expansion} could be an approximation rather than an equality when the number of dictionary functions is insufficient, we assume, for simplicity, that the equality holds in the following discussions. Applying the Koopman operator $\mathcal{K}$ to the observable function $\phi(\bm{x})$, we have
\begin{align}
\left(\mathcal{K}\phi\right)(\bm{x}_{i})  
= \sum_{n=1}^{N_{\mathrm{dic}}} a_{n}^{\phi} \mathcal{K} \psi_{n}(\bm{x}_{i}).
\end{align}
Note that $\{a_{n}^{\phi}\}_{n=1}^{N_{\mathrm{dic}}}$ is time-independent, and hence, it is enough to consider the action of $\mathcal{K}$ on the dictionary functions as follows:
\begin{align}
\bm{\psi}(\bm{x}_{i+1}) = \mathcal{K} \bm{\psi}(\bm{x}_{i}) \simeq K \bm{\psi}(\bm{x}_{i}),
\label{eq_Koopman_matrix}
\end{align}
which leads to the Koopman matrix $K$.

We here comment on the dictionary. There are many choices for dictionary functions, such as monomials, radial basis functions (RBFs), and Hermite polynomials. In the following discussions and numerical experiments, we employ monomial functions. For example, we consider a two-dimensional system with $\{x_1, x_2\}$ and the monomial functions with maximum degree $5$. Then, the dictionary $\bm{\psi}$ is given as
\begin{align}
\bm{\psi} (\bm{x})=
\begin{bmatrix}
1 &
x_1 &
x_2 &
x_{1}^{2} &
x_{1} x_{2} &
x_{2}^{2} &
\cdots &
x_{2}^{5}
\end{bmatrix}^{\top}.
\label{eq_monomial_dict}
\end{align}
In this case, the dictionary size is $N_{\mathrm{dic}}=21$.

Let $\Psi(X_{m})$ and $\Psi(Y_{m})$ be data matrices using the dictionary,
\begin{align}
\Psi(X_{m}) &=[\bm{\psi}(\bm{x}_{1})\ \bm{\psi}(\bm{x}_{2})\ \dots\ \bm{\psi}(\bm{x}_{m}) ], \\
\Psi(Y_{m}) &=[\bm{\psi}(\bm{y}_{1})\ \bm{\psi}(\bm{y}_{2})\ \dots\ \bm{\psi}(\bm{y}_{m}) ].
\end{align}
In the EDMD algorithm \cite{Williams2015}, we explore a solution that minimizes the following cost function:
\begin{align}
J(K_{m}) = \|\Psi(Y_{m}) - K_{m} \Psi(X_{m})\|_\mathrm{F}^2,
\label{eq_edmd_cost_function}
\end{align}
where $\| \cdot \|_\mathrm{F}$ is the Frobenius norm. The minimization problem for Eq.~\eqref{eq_edmd_cost_function} is a conventional least-squares problem for the dataset with the $m$ snapshot pairs, and then it is easy to obtain the solution $K_m$.

Finally, note that the monomial dictionary contains the monomial functions of the first degree. For example, the dictionary function $\psi_2(\bm{x}) = x_1$ directly yields the value of the state variable for the 1st dimension. Hence, we achieve the discrete time-evolution in Eq.~\eqref{eq_discrete_evolution} using the Koopman matrix $K$. That is, although the time-evolution in Eq.~\eqref{eq_discrete_evolution} would be nonlinear, only the linear matrix product is enough.

\section{Proposal for Forward Problem}
\label{sec_proposal_1}

\subsection{Overview of the proposed method}

\begin{figure}[b]
\centering
\includegraphics[width=70mm]{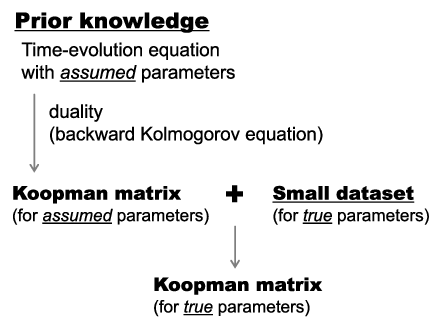}
\caption{Proposed method for the forward problem. The information for the underlying time-evolution equations is available, but the parameter values are unknown. The ambiguous prior knowledge leads to the prior Koopman matrix, and the update with a small dataset yields the approximated Koopman matrix for the true parameters.}
\label{fig_proposal_1}
\end{figure}

Here, we propose a method to construct the Koopman matrix from only a small dataset with the aid of ambiguous prior knowledge. Figure~\ref{fig_proposal_1} shows the proposed method for the forward problem, i.e., the prediction task. The ambiguous prior knowledge corresponds to the information for the underlying time-evolution equations. Note that the time-evolution equations have some parameters, and we do not know the parameter values. Using the ambiguous prior knowledge, we first construct the Koopman matrix for certain parameter values. The duality relation developed in Ref.~\cite{Ohkubo2022} yields the Koopman matrix. Next, the online EDMD algorithm \cite{Zhang2019} leads to the updated Koopman matrix with only a small dataset. Note that a naive approach to the online EDMD algorithm requires a dataset for constructing an initial Koopman matrix. There is no such dataset for initialization, and we will discuss this initialization procedure.

\subsection{Step 1: Initial Koopman matrix obtained from equation}

In the present paper, we consider only deterministic time-evolution equations, and we briefly reviewed the EDMD algorithm for the deterministic cases in Sect.~\ref{sec_review_Koopman}. It is possible to apply the EDMD algorithm to stochastic cases; for details, see Refs.~\cite{Williams2015,Crnjaric-Zic2020}. In addition, there have been many works on the duality relations in stochastic processes; for example, see the review paper in Ref.~\cite{Jansen2014}. The stochastic processes are connected naturally to the discussions on the Koopman operator in stochastic cases; the Fokker-Planck equation, which is related to the time-evolution of stochastic systems, has been connected the Koopman matrix \cite{Ohkubo2022}. The discussions for the stochastic cases are also available for our deterministic cases. Hence, it is straightforward to evaluate the Koopman matrix from the ambiguous prior knowledge if we choose a certain parameter set for the time-evolution equations.

The discussions to derive the Koopman matrix are complicated, and then we will explain them in Appendix~\ref{sec_duality}. In the following, we assume that the Koopman matrix has already been derived from the information of the equation with randomly chosen parameters.

\subsection{Step 2: Learning with a small dataset}

\begin{figure}[b]
\centering
\includegraphics[width=70mm]{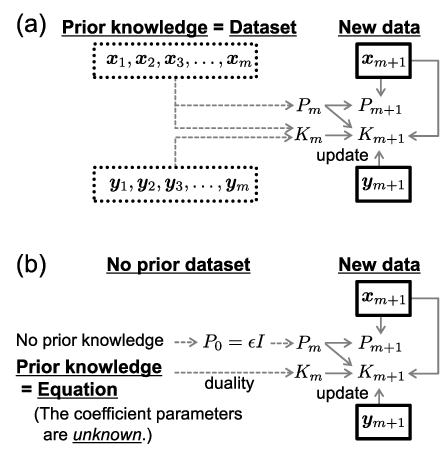}
\caption{(a) The conventional online EDMD algorithm. A prior dataset leads to the Koopman matrix $K_m$ and the intermediate matrix $P_m$, which are necessary for the update. (b) The proposed method. Since there is no prior dataset, the intermediate matrix $P_m$ is set as a simple identity matrix $I$ multiplied by a scalar parameter $\epsilon$. We apply the update procedure with the initial Koopman matrix $K_m$ constructed in Step 1.}
\label{fig_update_flow}
\end{figure}

We update the Koopman matrix constructed in Step 1 with a small dataset; the online EDMD algorithm leads to the update procedure. Figure~\ref{fig_update_flow}(a) shows the conventional online EDMD algorithm; for details, see Ref.~\cite{Zhang2019}. We here briefly explain the update procedure.

In the conventional online EDMD algorithm, the intermediate matrix $P_{m}$ is constructed using a prior dataset, i.e., $m$ snapshot pairs:
\begin{align}
P_{m} = \left(\Psi(X_{m}) \Psi(X_{m})^{\top}\right)^{-1}.
\end{align}
After obtaining the new snapshot pair, $(\bm{x}_{m+1}, \bm{y}_{m+1})$, the intermediate matrix $P_{m}$ is updated as follows:
\begin{align}
P_{m+1} = P_{m}
- \gamma_{m+1} P_{m} \bm{\psi}(\bm{x}_{m+1}) \bm{\psi}(\bm{x}_{m+1})^{\top} P_{m}, 
\label{eq_online_p_update}
\end{align}
where
\begin{align}
\gamma_{m+1} = \frac{1}{1 + \bm{\psi}(\bm{x}_{m+1})^{\top} P_{m} \bm{\psi}(\bm{x}_{m+1})}.
\end{align}
Then, the Koopman matrix is updated via
\begin{align}
K_{m+1} = K_{m}
+ \gamma_{m+1} \left(\bm{\psi}(\bm{y}_{m+1}) - K_{m} \bm{\psi}(\bm{x}_{m+1})\right) 
\bm{\psi}(\bm{x}_{m+1})^{\top} P_{m}.
\label{eq_online_k_update}
\end{align}
Note that there is no need to retain the snapshot pair, $(\bm{x}_{m+1}, \bm{y}_{m+1})$, after the update procedure.

In our problem settings, there is no dataset to construct the intermediate matrix $P_{m}$, as shown in Fig.~\ref{fig_update_flow}(b). Hence, we use the following matrix as the initial setting:
\begin{align}
P_m = \epsilon I,
\label{eq_P_epsilon}
\end{align}
i.e., a simple identity matrix $I$ multiplied by a scalar parameter $\epsilon$. As denoted above, the initial Koopman matrix $K_m$ is constructed from the duality relation. Then, the update procedure is repeated $M$ times for a small dataset with the size $M$, and we finally obtain the estimated Koopman matrix $\hat{K} \equiv K_{m+M}$. Although the initialization based on the identity matrix was commented on in Ref.~\cite{Zhang2019}, we will demonstrate that the combination of the initialization and the Koopman matrix from the ambiguous prior knowledge achieves the learning with only a small dataset.

We here comment on the role of the parameter $\epsilon$ which adjusts the weight of update data in the online EDMD procedures. A larger $\epsilon$ results in stronger adaptation to new datasets, whereas a smaller $\epsilon$ enforces stronger adherence to the initial Koopman matrix. We see this effect by substituting $P = \epsilon I$ into Eq.~\eqref{eq_online_k_update}:
\begin{align}
K_{m+1} = K_m
 + \epsilon \gamma_{m+1} \left(\bm{\psi}(\bm{y}_{m+1}) - K_{m} \bm{\psi}(\bm{x}_{m+1})\right) 
 \bm{\psi}(\bm{x}_{m+1})^{\top},
\end{align}
which indicates that $\epsilon$ acts as a weighting factor on the update term.

\subsection{Numerical experiments}

In the following, we perform numerical experiments for the proposed method using two nonlinear dynamical systems that have been used as examples in the previous works for the Koopman operators \cite{Williams2015,Crnjaric-Zic2020}. One is the Duffing equation, which exhibits convergence to a fixed point, and the other is the van der Pol equation, which converges to a limit cycle. The time-evolution equations are described as follows:
\begin{align}
\textrm{(Duffing)} \quad &
\begin{cases}
\,\, \dot{x}_1 = x_2, \\
\,\, \dot{x}_2 = -\delta x_2 - x_1(\beta + \alpha x_1^2),
\end{cases} \label{eq_duffing}\\
\textrm{(van der Pol)} \quad &
\begin{cases}
\,\, \dot{x}_1 = x_2, \\
\,\, \dot{x}_2 = -\mu (1-x_{1}^{2})x_{2}-x_{1},
\end{cases} \label{eq_van_der_pol}
\end{align}
where $\alpha, \beta$ and $\delta$ are the system parameters for the Duffing equation, and $\mu$ is that for the van der Pol equation.

In all of the following numerical experiments, we employ the monomial dictionary functions with maximum degree 5, as in Eq.~\eqref{eq_monomial_dict}.

\subsubsection{Duffing equation}
\label{sec_duffing_results}

For the Duffing equation in Eq.~\eqref{eq_duffing}, we set the following parameters as the true values: $\alpha_\mathrm{true} = 1.9$, $\beta_\mathrm{true} = -1.9$, and $\delta_\mathrm{true} = 1.4$. As described above, we cannot know these parameter values in advance, and it is necessary to construct the prior Koopman matrix $K_m$ using different parameter values. Since the learning is too easy when using parameters close to the true values, we set the assumed parameters differently from the true ones; $\alpha_\mathrm{assumed} = 1.0$, $\beta_\mathrm{assumed} = -1.0$, and $\delta_\mathrm{assumed} = 0.5$. Note that we tried several other parameters and confirmed that the proposed method works well.

The training datasets are generated as follows. First, we select the initial coordinates randomly in the range $[-1.0, 1.0]$ for both $x_1$ and $x_2$. Then, using the direct numerical integration with \verb|scipy.integrate.odeint| \cite{scipy}, we obtain $M$ snapshot pairs with the observation time $\Delta t_{\mathrm{obs}} = 0.1$. In the following numerical experiments, we consider the small dataset cases, and $M = 10$ is used. As for the parameter $\epsilon$ in Eq.~\eqref{eq_P_epsilon}, we set $\epsilon = 10^{10}$, which means stronger adaptations to the new dataset.

After the construction of the prior Koopman matrix, we apply the update with only the $M=10$ snapshot pairs. Using the estimated Koopman matrix $\hat{K}$, we evaluate the prediction errors, as follows. First, the initial states are sampled from a $5 \times 5$ grid in the range $[-1.0, 1.0]$ for $x_1$ and $x_2$, respectively. Next, similar to the generation of the training dataset, we make the ground truth trajectories by direct numerical integration with the same observation time $\Delta t_{\mathrm{obs}} = 0.1$. Then, we construct a matrix aligning the predicted trajectories and that for the true trajectories, respectively. Evaluating the Frobenius norm of the difference between the two matrices yields the prediction error.

\begin{figure}[bt]
\centering
\includegraphics[width=80mm]{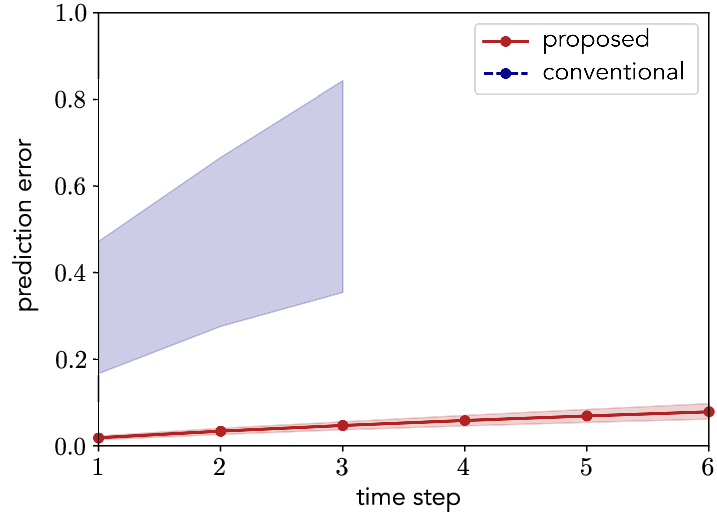}
\caption{(Color online) Prediction errors for the Duffing equation. The number of data is 10, and the prediction error was evaluated. We performed 100 trials. The solid line corresponds to the mean of the prediction errors of the proposed method, and the shaded area on the lower side represents the quartile range. The shaded area on the upper side represents the quartile range derived from the results of the conventional EDMD algorithm; since the mean of the prediction errors show divergent behavior immediately, we only draw it up to time step 3. We did not draw the dashed line representing the mean of the prediction errors because it goes beyond this drawing range and shows divergent behavior. }
\label{fig_duffing_prediction_results}
\end{figure}

We repeated the above prediction error estimation 100 times using different training datasets. Figure~\ref{fig_duffing_prediction_results} shows the prediction errors; the solid line corresponds to the mean of the prediction errors of the 100 trials, and the shaded area on the lower side represents the quartile range between the 25th and 75th percentiles. For comparison, the results of the conventional EDMD algorithm using $10$ data are also depicted; the mean of the prediction errors shows divergent behavior immediately, and only the quartile range is depicted. The relationship between the interquartile range and the mean of the prediction errors indicates that the prediction results of some trials are very poor, and we confirmed that the prediction was unstable in regions away from the training dataset. In contrast, the proposed method shows a small error with a narrow range. The reason is that the prior Koopman matrix contains the information for the time-evolution, even though the parameter values are different from the true ones.

\begin{figure}[bt]
\centering
\includegraphics[width=70mm]{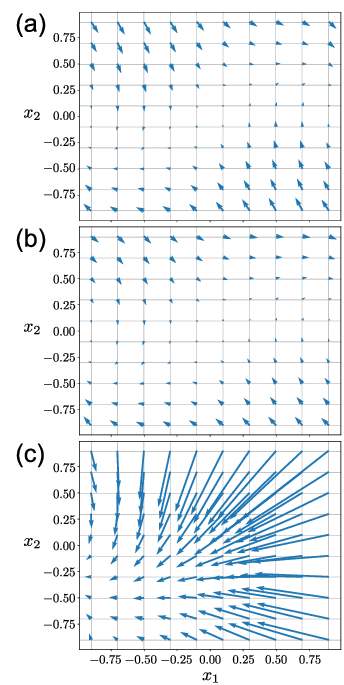}
\caption{(Color online) Examples of one-step predictions for the Duffing equation. We use each grid point as the initial coordinate. (a) True trajectories. (b) Trajectories by the proposed method. (c) Trajectories by the conventional EDMD algorithm.}
\label{fig_duffing_one_stap_predictions}
\end{figure}

Next, we show examples of one-step predictions in Fig.~\ref{fig_duffing_one_stap_predictions}. The initial coordinates are selected on the grid point, as in the prediction error estimations. Figure~\ref{fig_duffing_one_stap_predictions}(a) shows the true trajectories via the direct numerical integration; Fig.~\ref{fig_duffing_one_stap_predictions}(b) and Fig.~\ref{fig_duffing_one_stap_predictions}(c) correspond to the results obtained from the proposed method and the conventional EDMD algorithm, respectively. As we see from Fig.~\ref{fig_duffing_one_stap_predictions}, the conventional EDMD algorithm yields a large deviation from the ground truth, especially in the right half of the domain; actually, the training dataset was located in the left region in this example. By contrast, the proposed method leads to stable and qualitatively accurate predictions across the whole domain. The results indicate that even ambiguous prior knowledge is useful enough. Note that the results in Figs.~\ref{fig_duffing_one_stap_predictions}(b) and (c) depend on the learning datasets, and we confirmed that the final consequences do not differ when the datasets change.

\subsubsection{Van der Pol equation}
\label{sec_van_der_pol_results}

\begin{figure}[bt]
\centering
\includegraphics[width=80mm]{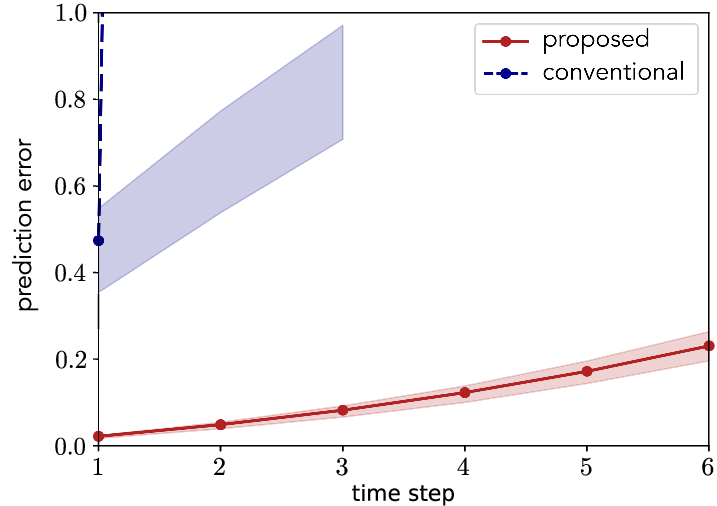}
\caption{(Color online) Prediction errors for the van der Pol equation. The number of data is 10. The solid line corresponds to the mean of the prediction errors of the proposed method, and the dashed line is the result of the conventional EDMD algorithm. The shaded areas correspond to the interquartile ranges. As in Fig.~\ref{fig_duffing_prediction_results}, the conventional EDMD yielded divergent behavior, and then, we only draw it up to time step 3.
}
\label{fig_van_der_pol_prediction_results}
\end{figure}

\begin{figure}[bt]
\centering
\includegraphics[width=70mm]{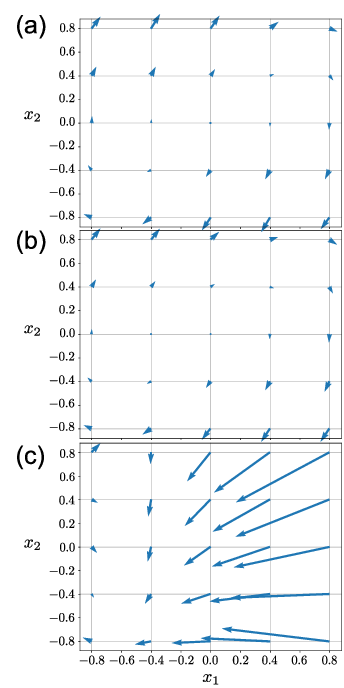}
\caption{(Color online) Examples of one-step predictions for the van der Pol equation. We use each grid point as the initial coordinate. (a) True trajectories. (b) Trajectories by the proposed method. (c) Trajectories by the conventional EDMD algorithm.}
\label{fig_van_der_pol_one_stap_predictions}
\end{figure}

We here show the numerical results for the van der Pol equation in Eq.~\eqref{eq_van_der_pol}. We use $\mu_\mathrm{true} = 1.9$ as the true parameter value; the prior Koopman matrix $K_m$ is constructed with $\mu_\mathrm{assumed} = 1.0$. Other settings are the same with Sect.~\ref{sec_duffing_results}.

Figures~\ref{fig_van_der_pol_prediction_results} and \ref{fig_van_der_pol_one_stap_predictions} show the prediction errors and the one-step predictions, respectively. The van der Pol system rapidly converges to a limit cycle from various initial conditions, and the small dataset with $M=10$ could prevent us from capturing the whole dynamics. Nevertheless, we see that the proposed method captures the overall trend of the dynamics well compared with the conventional EDMD algorithm.

\section{Proposal for Inverse Problems}
\label{sec_proposal_2}

In this section, we propose two approaches to tackle the inverse problem of estimating the parameters of the time-evolution equations. One is an extension of the proposed method in Sect.~\ref{sec_proposal_1}, and the other is based on optimization procedures.

\subsection{Approach 1: Extension of the small dataset learning}
\label{sec_inverse_approach_1}

\subsubsection{Overview of approach 1 for the inverse problems}

\begin{figure}[bt]
\centering
\includegraphics[width=70mm]{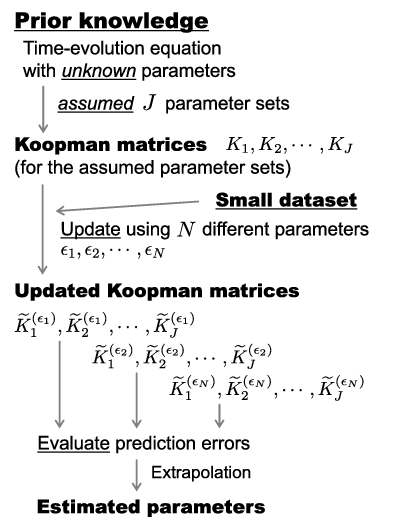}
\caption{Proposed method for the inverse problem. We first assume $J$ parameter sets and construct the prior Koopman matrix for each parameter set. Next, we evaluate the prediction errors for each Koopman matrix using different $\epsilon$ values in the update. Then, the extrapolations with the relationship between the prediction error and the parameters lead to the estimations for the true parameters.}
\label{fig_proposal_2}
\end{figure}

Figure~\ref{fig_proposal_2} shows the concept for the first approach. In this approach, we construct various prior Koopman matrices using various parameters. Here, $J$ parameter sets are assumed, which lead to $J$ prior Koopman matrices. Next, the update procedure in Sect.~\ref{sec_proposal_1} is applied with $M$ snapshot pairs. Note that the parameter $\epsilon$ in Eq.~\eqref{eq_P_epsilon} affects the estimation, and we choose $N$ different values for $\epsilon$. Using the various parameter $\epsilon$, we have various estimated Koopman matrices, as shown in Fig.~\ref{fig_proposal_2}. Then, the prediction errors for each estimated Koopman matrix are evaluated, and finally the extrapolations with the relationship between the prediction error and the parameters lead to the estimations for the true parameters. 

Since it would be better to demonstrate the approach using concrete examples, see the numerical results for the van der Pol equation below.

\subsubsection{Numerical results for the van der Pol equation}

\begin{figure}[bt]
\centering
\includegraphics[width=80mm]{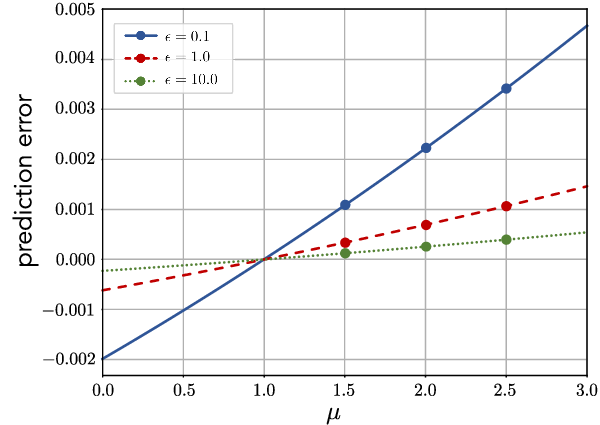}
\caption{(Color online) Prediction errors with various $\mu_{\mathrm{assumed},j}$ and $\epsilon_n$. The solid, dashed, and dotted lines are the fitting results by quadratic functions for $\epsilon_1 = 0.1$, $\epsilon_2 = 1.0$, and $\epsilon_3 = 10.0$ cases, respectively. Note that the prediction errors are evaluated from the estimated trajectories of the final estimated Koopman matrix and the observed snapshot pairs used in the update.}
\label{fig_estimation_resutls_for_van_der_Pol}
\end{figure}

The van der Pol equation in Eq.~\eqref{eq_van_der_pol} has only one system parameter, $\mu$, and hence it is easy to see the procedure. Here, we set $\mu_\mathrm{true} = 1.0$ as the true parameter value. The data generation method is the same as in Sect.~\ref{sec_proposal_1}; we use only $M=10$ snapshot pairs.

First, we set several assumed parameters; here, $\mu_{\mathrm{assumed},1} = 1.5$, $\mu_{\mathrm{assumed},2} = 2.0$, and $\mu_{\mathrm{assumed},3} = 2.5$ are used. Second, the prior Koopman matrix is generated as in the case of Sect.~\ref{sec_proposal_1} for each assumed parameter $\mu_{\mathrm{assumed},j}$. Third, the same update procedure as in Sect.~\ref{sec_proposal_1} is performed, using several parameters $\epsilon$; we here use $\epsilon_1 = 0.1$, $\epsilon_2 = 1.0$, and $\epsilon_3 = 10.0$. Note that, different from Sect.~\ref{sec_proposal_1}, we use small $\epsilon$ values to reflect the nature of the prior Koopman matrix.

Figure~\ref{fig_estimation_resutls_for_van_der_Pol} shows the prediction errors for various settings; the solid, dashed, and dotted lines are the fitting results by quadratic functions for $\epsilon_1 = 0.1$, $\epsilon_2 = 1.0$, and $\epsilon_3 = 10.0$ cases, respectively. The prediction errors are evaluated from the estimated trajectories of the final estimated Koopman matrix and the observed snapshot pairs used in the update. This is because we aim to parameter estimation tasks on small amounts of data. Here, we use only $M=10$ snapshot pairs for the training and the evaluation of errors.

From Fig.~\ref{fig_estimation_resutls_for_van_der_Pol}, we see that the fitting results calculated for the various parameters $\{\epsilon_n\}$ intersect at a single point, i.e., $\hat{\mu} = 0.999$. This intersection point corresponds to the estimated value, which is in close agreement with the true value $\mu_\mathrm{true} = 1.0$. In addition, this intersection point has the prediction error value of almost zero. From this reason, we can say that this intersection point gives the estimate.

Why does the intersection point yield the estimated parameter value? Of course, one of the reasons is that the point with the zero prediction error should be common. There would be another reason, as follows. If we use the true parameter value as the initial one in the proposed method, the prior knowledge exactly corresponds to the dataset. Hence, the update procedure will have essentially no effect. Since the parameter $\epsilon$ affects the update, the intersection point means the update has no effect. Therefore, the intersection yields an estimation for the parameter.

\subsection{Approach 2: Method based on optimization procedures}

\subsubsection{Overview of approach 2 for the inverse problems}

When one estimates multiple parameters, there could be no intersection for fitting results in parameter regions far from the true ones. Hence, we need iterative computations. Due to the computational complexity and instability issues in such cases, we here propose another approach based on a simple optimization algorithm. There are many optimization algorithms \cite{Forst_book}, and we here employ the Nelder-Mead optimization algorithm.

The procedures in this approach are as follows:
\begin{enumerate}
\item \textbf{Initialization}: Randomly select initial parameter values within a certain parameter region.
\item \textbf{Koopman matrix construction}: Construct the Koopman matrix for the candidate parameters using the method based on the duality relation.
\item \textbf{Time-evolution prediction}: Compute the prediction error between the observed snapshot pairs and the snapshot pairs obtained from the Koopman matrix.
\item \textbf{Optimization}: Apply the Nelder-Mead method to iteratively refine the parameters, minimizing the prediction error. Note that the Koopman matrix is recomputed with the new parameter values in each iteration step.
\end{enumerate}
Note that this approach does not include the procedure for updating the Koopman matrix with a small dataset. Instead, the prediction error is evaluated on the small dataset. In addition, this approach leverages the efficiency of the Nelder-Mead algorithm, enabling parameter space exploration with fewer evaluations compared to exhaustive methods. Incorporating the ambiguous prior knowledge with the prior Koopman matrices ensures robust estimation even with the small dataset, as we will see in the next numerical demonstration.

\subsubsection{Numerical results for the Duffing equation}

The Duffing equation in Eq.~\eqref{eq_duffing} has three parameters, $\alpha$, $\beta$, and $\delta$. We here set the true parameter values, $\alpha_\mathrm{true}$, $\beta_\mathrm{true}$, and $\delta_\mathrm{true}$, randomly; each parameter is chosen from the $[-10,10]$ region, respectively. In addition, we estimate the parameters only with $M=10$ snapshot pairs. The data generation method is the same as in Sect.~\ref{sec_proposal_1}.

In the numerical experiments, we randomly choose each initial parameter from the $[-10,10]$ region, respectively. Then, according to the above procedures, we search for the optimum estimations using the Nelder-Mead algorithm. To assess the accuracy of parameter estimation, we conduct $10$ independent trials with different random true parameters and initial guesses. The estimated parameters are compared with the true values, and the total error for the three parameters is evaluated in terms of the vector norm between the true parameter values and the estimated values.

As a result, the mean estimation error was $0.005431 \pm 0.005289$; this error is sufficiently small even though we used only $10$ snapshot pairs. In addition, the small standard deviation indicates stable estimation across trials. The computational time was $16.43 \pm 3.53$ seconds. Although we did not devise a way to speed up the computation, the estimation was completed in the practical computation time. 

Finally, we would like to comment on the optimization approach. The optimization approach is essentially the same as solving the nonlinear time-evolution equations for the initial values of all snapshot pairs. Note that we require the re-evaluation of the time-evolution many times during the parameter changes in the optimization process. These direct time-integration for the all initial conditions could be time-consuming. On the other hand, once the Koopman matrix is obtained for a certain parameter set, it is possible to evaluate the time-evolution for all initial values using only the matrix product; see the term $K_m \Psi(X_m)$ in Eq.~\eqref{eq_edmd_cost_function}. This linearity is a characteristic of the proposed method.

\section{Conclusion}

In this work, we proposed an integrated approach combining data-driven and equation-based techniques within the Koopman operator framework. Our target is deterministic time-evolution equations and a small amount of data. For the forward problems, i.e., the prediction tasks, we utilized a prior Koopman matrix, which is derived from ambiguous prior knowledge, with the online EDMD algorithm. We numerically demonstrated that the proposed method achieves stable and accurate time-evolution predictions even with the limited data points; the number of data was only $10$. For the inverse problems, we proposed two approaches; one is the extension of the algorithm for the forward problem, and the other is based on the Nelder-Mead optimization algorithm used for multiple parameter cases. It is possible to vary the parameter $\epsilon$ for the online updates, and we observed that the intersection occurs when we tried several parameters. Including the multiple parameter cases, we confirmed that the proposed methods work well even in the small dataset cases.

This work is the first trial to incorporate the information for time-evolution equations with data in the Koopman operator approach, and there are some remaining tasks in the future. The first one is to investigate the noise robustness. Although the incorporation of ambiguous prior knowledge could improve the accuracy of the prediction tasks, the presence of noise would have a significant impact on the inverse problem. The inverse problem, i.e., the parameter estimation task with only small amounts of data, needs further investigation. The work in Ref.~\cite{Tahara2024} aims to reduce the noise effects in the context of equation estimation using Koopman operators, and it will be a challenging task to solve inverse problems under noisy conditions using only small amounts of data. In such cases, the use of prior knowledge would be helpful, even if it has some ambiguities. The other remaining task is the applications in higher dimensional systems. It is known that the computation of the Koopman matrix from equations becomes computationally difficult for high-dimensional cases. Several attempts to solve this problem have been in progress \cite{Ohkubo2022,Kinjo2025}, and it would be valuable to seek the applications of the proposed method in this work to higher-dimensional systems in the future. We hope that this work will establish a versatile framework that bridges forward and inverse problems, paving the way for broader applications in nonlinear system analysis.

\begin{acknowledgments}
This work was supported by JST FOREST Program (Grant Number JPMJFR216K, Japan).
\end{acknowledgments}

\appendix
\section{Evaluate Koopman matrix from equations}
\label{sec_duality}

Let $\mathcal{M} \subseteq \mathbb{R}^D$ be the state space. We here consider the following time-evolution equation for a state vector $\bm{x} \in \mathcal{M}$:
\begin{align}
\frac{d}{dt} \bm{x}(t) = \bm{f}(\bm{x}(t)),
\label{eq_appendix_time_evolution}
\end{align}
where $\bm{f}(\bm{x})$ is a nonlinear function. Although we wrote explicitly the system parameters $\bm{\theta}$ in $\bm{f}$ in Eq.~\eqref{eq_time_evolution}, we omit them in the following discussion for notational simplicity. Then, the generator of the Perron-Frobenius operator is given by \cite{Klus2020}
\begin{align}
\mathcal{L} 
= - \sum_{d=1}^{D} \frac{\partial}{\partial x_d} f_d(\bm{x}),
\end{align}
which acts on an observable function $\phi: \mathcal{M} \to \mathcal{M}$. Note that $\mathcal{L}$ corresponds to the Fokker-Planck operator in the presence of diffusion terms, and then the discussions for the duality relations in stochastic processes in Ref.~\cite{Ohkubo2022} are straightforwardly applicable. As a result, it is possible to consider the time-evolution of the observable function $\phi$, instead of the state vector $\bm{x}$, as follows:
\begin{align}
\phi(\bm{x}(t)) &= \int_{\mathcal{M}} \phi(\bm{x}) e^{\mathcal{L}t} \delta\left(\bm{x} - \bm{x}(0)\right) d\bm{x}\nonumber \\
&= \int_{\mathcal{M}} \left(e^{\mathcal{L}^{\dagger} t} \phi(\bm{x}) \right)
 \delta\left(\bm{x}(t) - \bm{x}(0)\right) d\bm{x} \nonumber \\
&= \int_{\mathcal{M}} \phi(\bm{x},t)
 \delta\left(\bm{x}(t) - \bm{x}(0)\right) d\bm{x} \nonumber \\
&=  \phi(\bm{x}(0),t),
\label{eq_appendix_duality}
\end{align}
where $\delta(\cdot)$ is the Dirac delta function, and $\mathcal{L}^{\dagger}$ is the adjoint operator of $\mathcal{L}$ and is defined as
\begin{align}
\mathcal{L}^{\dagger} = \sum_{d=1}^{D}  f_d(\bm{x}) \frac{\partial}{\partial x_d}.
\label{eq_appendix_adjoint_op}
\end{align}
Note that the function $\phi(\bm{x},t)$ obeys the following time-evolution equation,
\begin{align}
\frac{d}{dt} \phi(\bm{x},t) = \mathcal{L}^\dagger \phi(\bm{x},t),
\label{eq_appendix_dual_time_evolution}
\end{align}
and the initial condition is the observable function $\phi(\bm{x})$, i.e.,
\begin{align}
\phi(\bm{x},t=0) = \phi(\bm{x}).
\end{align}
Then, the function $\phi(\bm{x}(0),t)$ immediately yields the value of the observable function $\phi(\bm{x})$ at time $t$.

In the following discussions, we focus on a monomial function $\psi_{\bm{\zeta}}(\bm{x})$,
\begin{align}
\psi_{\bm{\zeta}}(\bm{x}) &= \bm{x}^{\bm{\zeta}} 
\equiv x_1^{\zeta_1} x_2^{\zeta_2} \dots x_D^{\zeta_D},
\end{align}
where $\bm{\zeta} = (\zeta_1,\zeta_2,\dots,\zeta_D) \in \mathbb{N}_0^{D}$. Since Eq.~\eqref{eq_appendix_duality} leads to the observable function at time $t$, we expand it in terms of the monomial functions as follows:
\begin{align}
\psi_{\bm{\zeta}}(\bm{x},t)
= \sum_{\bm{\zeta}'} c^{\bm{\zeta}}({\bm{\zeta}'},t) x_1^{\zeta'_1} x_2^{\zeta'_2} \dots x_D^{\zeta'_D},
\label{eq_appendix_monomial_expansion}
\end{align}
where $c^{\bm{\zeta}}({\bm{\zeta}'},t)$ is a time-dependent expansion coefficient of the term $\bm{x}^{\bm{\zeta}'}$ for the function $\psi_{\bm{\zeta}}(\bm{x},t)$. Although we need an approximation with a finite number of monomials, Eq.~\eqref{eq_appendix_monomial_expansion} expresses the time-evolved monomial function, $\psi_{\bm{\zeta}}(\bm{x},t)$, as a linear combination of monomial functions. Then, we see that Eq.~\eqref{eq_appendix_monomial_expansion} directly corresponds to the relation in Eq.~\eqref{eq_Koopman_matrix}. Hence, the elements of the Koopman matrix are $\{c^{\bm{\zeta}}({\bm{\zeta}'},t)\}$.

The remaining task is to connect the time-evolution equation in Eq.~\eqref{eq_appendix_dual_time_evolution} with the time-evolution equations for $\{c^{\bm{\zeta}}({\bm{\zeta}'},t)\}$. The basis expansion with the monomial functions leads to the connection, and we here yield a demonstration using a concrete example, i.e., the Duffing equation in Eq.~\eqref{eq_duffing}. As for the Duffing equation, the adjoint operator $\mathcal{L}^\dagger$ in Eq.~\eqref{eq_appendix_adjoint_op} is
\begin{align}
\mathcal{L}^{\dagger} 
= x_2 \frac{\partial}{\partial x_1} 
+ \left(-\delta x_2 - x_1(\beta + \alpha x_1^2)\right) \frac{\partial}{\partial x_2},
\end{align}
and we should solve the following partial differential equation:
\begin{align}
\frac{\partial}{\partial t} \psi_{\bm{\zeta}}(\bm{x}, t) 
= x_2 \frac{\partial}{\partial x_1} \psi_{\bm{\zeta}}(\bm{x}, t) 
+ \left(-\delta x_2 - x_1(\beta + \alpha x_1^2)\right) \frac{\partial}{\partial x_2}
\psi_{\bm{\zeta}}(\bm{x}, t).
\label{eq_appendix_pde}
\end{align}

By applying the basis expansion in Eq.~\eqref{eq_appendix_monomial_expansion}, we have
\begin{align}
&  \sum_{\zeta'_1, \zeta'_2} \frac{\partial}{\partial t}
c^{\bm{\zeta}}(\zeta'_1,\zeta'_2,t)
x_1^{\zeta'_1} x_2^{\zeta'_2} \nonumber \\
&= \sum_{\zeta'_1, \zeta'_2} 
\Big(
(\zeta'_1 + 1) c^{\bm{\zeta}}(\zeta'_1 + 1, \zeta'_2 - 1, t) \nonumber \\
&\qquad \quad + \delta \zeta'_2 c^{\bm{\zeta}}(\zeta'_1, \zeta'_2, t) \nonumber \\
&\qquad \quad - \beta(\zeta'_2 - 1) c^{\bm{\zeta}}(\zeta'_1 -1, \zeta'_2 + 1, t) \nonumber \\
&\qquad \quad - \alpha(\zeta'_2 + 1) c^{\bm{\zeta}}(\zeta'_1 -3, \zeta'_2 + 1, t) 
\Big) x_1^{\zeta'_1} x_2^{\zeta'_2}.
\end{align}
The comparison of coefficients for $x_1^{\zeta'_1}x_2^{\zeta'_2}$ on both sides leads to
\begin{align}
\frac{\partial}{\partial t}
c^{\bm{\zeta}}(\zeta'_1,\zeta'_2,t) 
=&(\zeta'_1 + 1) c^{\bm{\zeta}}(\zeta'_1 + 1, \zeta'_2 - 1, t) \nonumber \\
& + \delta \zeta'_2 c^{\bm{\zeta}}(\zeta'_1, \zeta'_2, t) \nonumber \\
& - \beta(\zeta'_2 - 1) c^{\bm{\zeta}}(\zeta'_1 -1, \zeta'_2 + 1, t) \nonumber \\
& - \alpha(\zeta'_2 + 1) c^{\bm{\zeta}}(\zeta'_1 -3, \zeta'_2 + 1, t).
\label{eq_appendix_final_coupled_odes}
\end{align}
Then, it is enough to numerically solve the coupled ordinary differential equations in Eq.~\eqref{eq_appendix_final_coupled_odes}. The initial condition for $\{c^{\bm{\zeta}}(\zeta'_1,\zeta'_2,t)\}$ corresponds to which column of the Koopman matrix is computed. If we compute the column of a dictionary function with $\zeta_1 = a$ and $\zeta_2 = b$, the following initial condition is used:
\begin{align}
c^{\bm{\zeta}}(\zeta'_1,\zeta'_2,t=0)
=
\begin{cases}
\, 1 & (\zeta'_1 = a, \zeta'_2 = b), \\
\, 0 & \text{otherwise},
\end{cases}
\label{eq_appendix_initial_condition}
\end{align}
because the initial condition leads to
\begin{align}
\psi_{\zeta_1=a,\zeta_2=b}(\bm{x},t=0)
= \psi_{\zeta_1=a,\zeta_2=b}(\bm{x}) 
= x_1^{a} x_2^{b},
\end{align}
which corresponds to the evaluation of the time-evolution of the monomial function $x_1^{a} x_2^{b}$.

In this work, we consider the monomial functions with maximum degree $5$. Hence, the number of the coupled ordinary differential equations is $21$. Solving the $21$ coupled ordinary differential equations with a specific initial condition in Eq.~\eqref{eq_appendix_initial_condition} yields a column of the Koopman matrix. Repeating this procedure for all initial conditions constructs the full Koopman matrix.

\end{document}